\documentclass[runningheads]{llncs}
\usepackage{graphicx}
\usepackage{multirow}
\usepackage{multicol}
\usepackage{subcaption}
\usepackage{enumitem}
\usepackage{xcolor}
\usepackage{arydshln}

\def\barr{\begin{tabular}{l}}
\def\earr{\end{tabular}}
\begin{document}

\title{Semi-supervised Learning of Fetal Anatomy from Ultrasound}

%
\titlerunning{Semi-supervised Learning for Fetal Ultrasound}
%
\author{Jeremy Tan \and
Anselm Au \and
Qingjie Meng\and
Bernhard Kainz}
\authorrunning{Tan J. et al.}
%
\institute{Imperial College London, SW7 2AZ, London, UK \\
\email{j.tan17@imperial.ac.uk}\\
}

%
\maketitle              
\begin{abstract}
Semi-supervised learning methods have achieved excellent performance on standard benchmark datasets using very few labelled images. Anatomy classification in fetal 2D ultrasound is an ideal problem setting to test whether these results translate to \textit{non}-ideal data. Our results indicate that inclusion of a challenging background class can be detrimental and that semi-supervised learning mostly benefits classes that are already distinct, sometimes at the expense of more similar classes. 

\keywords{Semi-supervised Learning  \and Fetal Ultrasound}%
\end{abstract}%

\section{Introduction}
Fetal ultrasound is the most widespread screening tool for congenital abnormalities and is a key recommendation in the World Health Organization's guidelines for antenatal care~\cite{WHO2016recommendations}. Classification of standardized tomographic 2D planes is a key step in nearly all screening exams. However, low image quality and shortage of experts can compromise screening efficacy or in the least make quality heterogeneous across sites. To democratize care, several efforts have been made to automate standard plane detection using deep learning~\cite{Baumgartner2017SonoNet,cai2018multi,Chen2015PlaneDetectionDomainTransfer,Kong2018PlaneDetectionMultiScale}. However, many of these methods still rely on large amounts of labelled data.

Because labelling is expensive, semi-supervised learning (SSL) methods have become an active area of research, particularly for image data~\cite{Laine2017Pi,Xie2019UDA} and in the medical domain~\cite{Cheplygina2019NotSoSupervised}. Recent methods have achieved remarkable performance by learning from unlabelled data. This added information can help to push decision boundaries into lower density regions, resulting in better generalization. Amidst this progress, Oliver et al. call for more ``realistic evaluation''~\cite{Oliver2018RealisticEvaluation}. One key concern is that benchmark datasets (e.g. CIFAR10, SVHN) do not reflect realistic scenarios. 

We study the use of SSL for standard plane classification in fetal ultrasound~\cite{Baumgartner2017SonoNet}. This task includes a challenging background class, class imbalance, and different levels of inter-class similarity. In accordance with Oliver et al. we demonstrate that supervised baselines can cope with surprisingly few labelled images; that a background class can cause SSL to become detrimental; and that SSL is effective for distinct classes, but can weaken performance on classes which are prone to confusion. 

\section{Related Work}
Automatic anatomy detection from ultrasound videos is a popular topic with many successful approaches using convolutional neural networks~\cite{Baumgartner2017SonoNet}. Further advancements have added multi-task learning to predict sonographer gaze~\cite{cai2018multi} or multi-scale networks to exploit lower and higher level features~\cite{Kong2018PlaneDetectionMultiScale}. Some methods have also studied the challenge of limited labelled data by pretraining on natural images~\cite{Chen2015PlaneDetectionDomainTransfer}. However there has been little investigation into the use of SSL. Exploring this avenue can reveal what benefits SSL can bring, and conversely, what challenges remain for SSL in non-ideal data scenarios. 

Recently, SSL methods have gained momentum.
Underlying many of these methods is a consistency loss which minimizes sensitivity to perturbations (in input/weight space). Examples of perturbations include image augmentations, gaussian noise, weight dropout~\cite{Laine2017Pi}, targeted augmentations~\cite{Xie2019UDA}, and mixup augmentations (interpolation between images)~\cite{Verma2019ICT}. Input perturbations have been shown to minimize the input-output Jacobian (linked to better generalization)~\cite{Athiwaratkun2019FastSWA}. Despite these advances, Oliver et al. point out that real unlabelled data is likely to be more irregular than the perfectly balanced benchmark datasets. It may also include out-of-distribution or confounding data that could hurt SSL \cite{Singh2008HelpDoesnt}. We aim to study the ways in which SSL might help in a real problem which stands to benefit from SSL.

\section{Methods}
The architecture used in this study, Sononet~\cite{Baumgartner2017SonoNet}, is a convolutional neural network (similar to VGG) that has been tailored for the task of anatomical standard plane detection in fetal ultrasound. It contains 15 convolutional layers, 4 maxpooling layers, and ends with global average pooling. This acts as a strong fully supervised baseline. Supervised methods are trained using Adam
with a learning rate of 1E-3. All models are trained for 50 epochs with a batch size of 32.

For the SSL method, we use the consistency loss employed in both the $\Pi$ model~\cite{Laine2017Pi} and the unsupervised data augmentation (UDA) method~\cite{Xie2019UDA} which are among the state of the art for standard benchmark datasets. This consistency loss uses the softmax predictions for unlabelled data, $\mathcal{D}_{U}$, as labels for the same images under augmentations. This consistency loss can be formalized as
\begin{equation}
\mathcal{L}_{\textrm{KL}}(x_u,w) = KL(f(x_u;w)||f(x'_u;w)).
\label{equation:KL_loss}
\end{equation}

This is added to the typical supervised (cross-entropy) loss with some proportion $\lambda$ (in all our experiments $\lambda = 0.5$). Note that the weights $w$ are the same for inference on both the original image $x_u$ (drawn from $\mathcal{D}_{U}$) and the corresponding augmented image $x'_u$. In this case the augmentations include random combinations of the following operations:
\begin{itemize}
	\item Horizontal flipping
	\item Random contrast adjustment by a factor within [0.7, 1.3]
	\item Random rotation by an angle within [$-\frac{\pi}{4}$, $\frac{\pi}{4}$]
	\item Random cropping ranging from 1\% to 20\%
\end{itemize}

The same augmentation is applied to all methods, including supervised baselines. These augmentations are not specially tailored to the data as is the case in UDA. For the best performance, UDA uses AutoAugment~\cite{Cubuk2019AutoAugment}, a reinforcement learning method that finds the optimal augmentation policies. 
This would likely improve both SSL and fully supervised regimes. However, as shown in the results, the scores of the supervised baselines are already very high with surprisingly few labelled images; applying AutoAugment to both SSL and fully supervised methods may truncate the margin for potential improvement that we wish to study. We include 
a) training signal annealing (TSA), b) confidence-based masking (CBM), c) entropy minimization, and d) softmax temperature controlling~\cite{Xie2019UDA}. Since these are proposed in \cite{Xie2019UDA}, we refer to the combination of i) the consistency loss with ii) these additional techniques, as UDA for simplicity and to acknowledge their contributions.

TSA~\cite{Xie2019UDA} uses a threshold, $\eta_{tsa}$, to mask the contribution of a given labelled image, $x_l$ (drawn from $\mathcal{D}_{L}$), to the supervised gradient. Specifically, an example only contributes to the gradient if the softmax probability in the ground truth class is greater than the threshold, $p(y^*|x_l) > \eta_{tsa}$. The threshold $\eta_{tsa}$ starts at $\frac{1}{\# of classes}$ and is increased to 1 following a linear, log, or exponential schedule across the total number of epochs.

CBM~\cite{Xie2019UDA} uses a confidence threshold on unlabelled images. An unlabelled image only contributes to the consistency gradient if, $\textrm{max}(p(y|x_u))>\eta_{cbm}$. A $\eta_{cbm}$ of 0.75 is used for all SSL experiments.

Entropy minimization~\cite{Grandvalet2005EntMin} is applied to $p(y|x'_u)$, the prediction for an augmented unlabelled image. Also the prediction for original unlabelled image is sharpened by using a softmax temperature of 0.8.

\begin{figure*}[h]
	\centering
	\begin{subfigure}[t]{0.48\linewidth}
	
        \includegraphics[width=\linewidth]{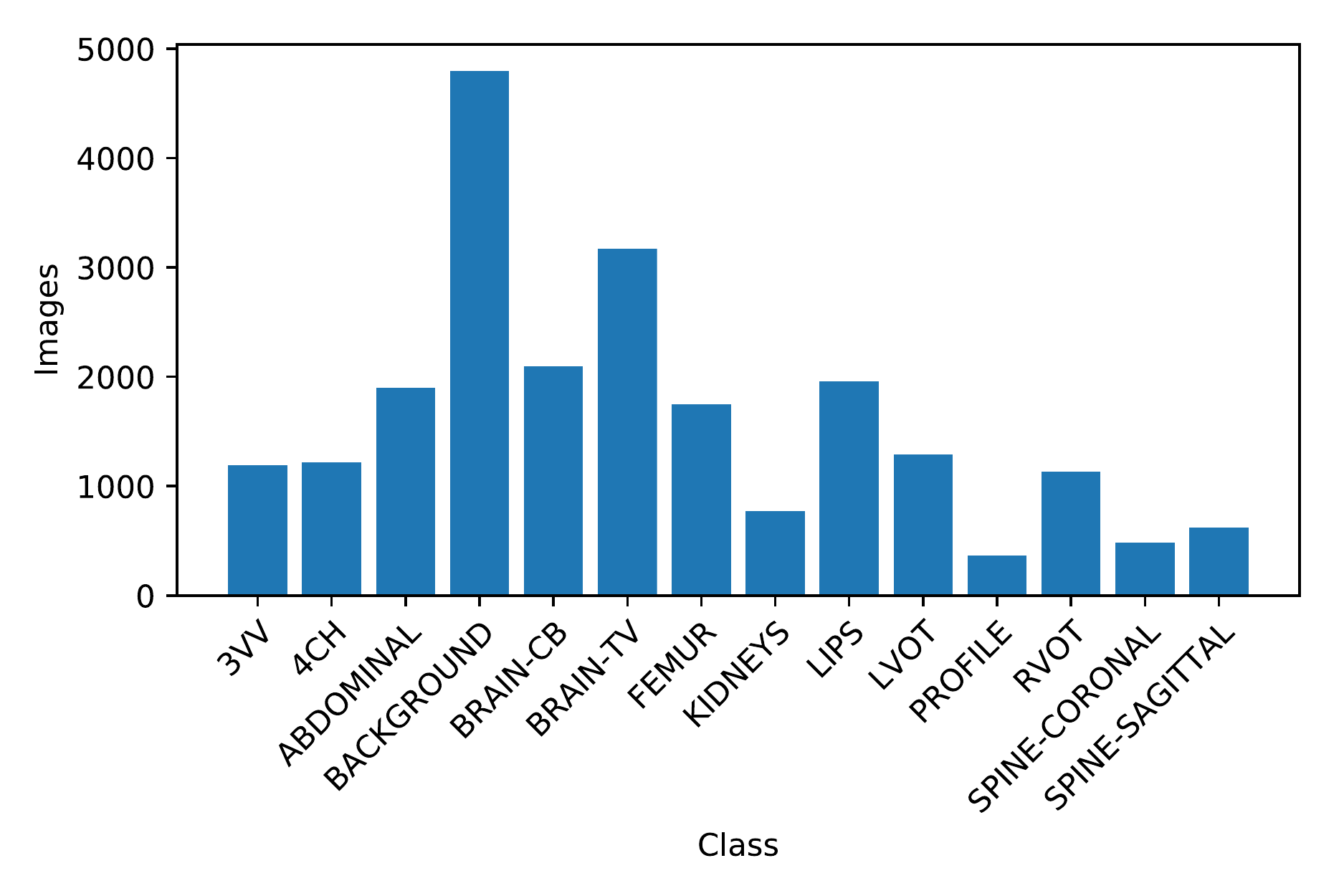}
		\caption{\textit{Data Distribution}}
	\end{subfigure}
	\hspace*{\fill}
	\begin{subfigure}[t]{0.48\linewidth}
		\includegraphics[trim={0cm 0cm 0cm 0cm},clip,width=\linewidth]{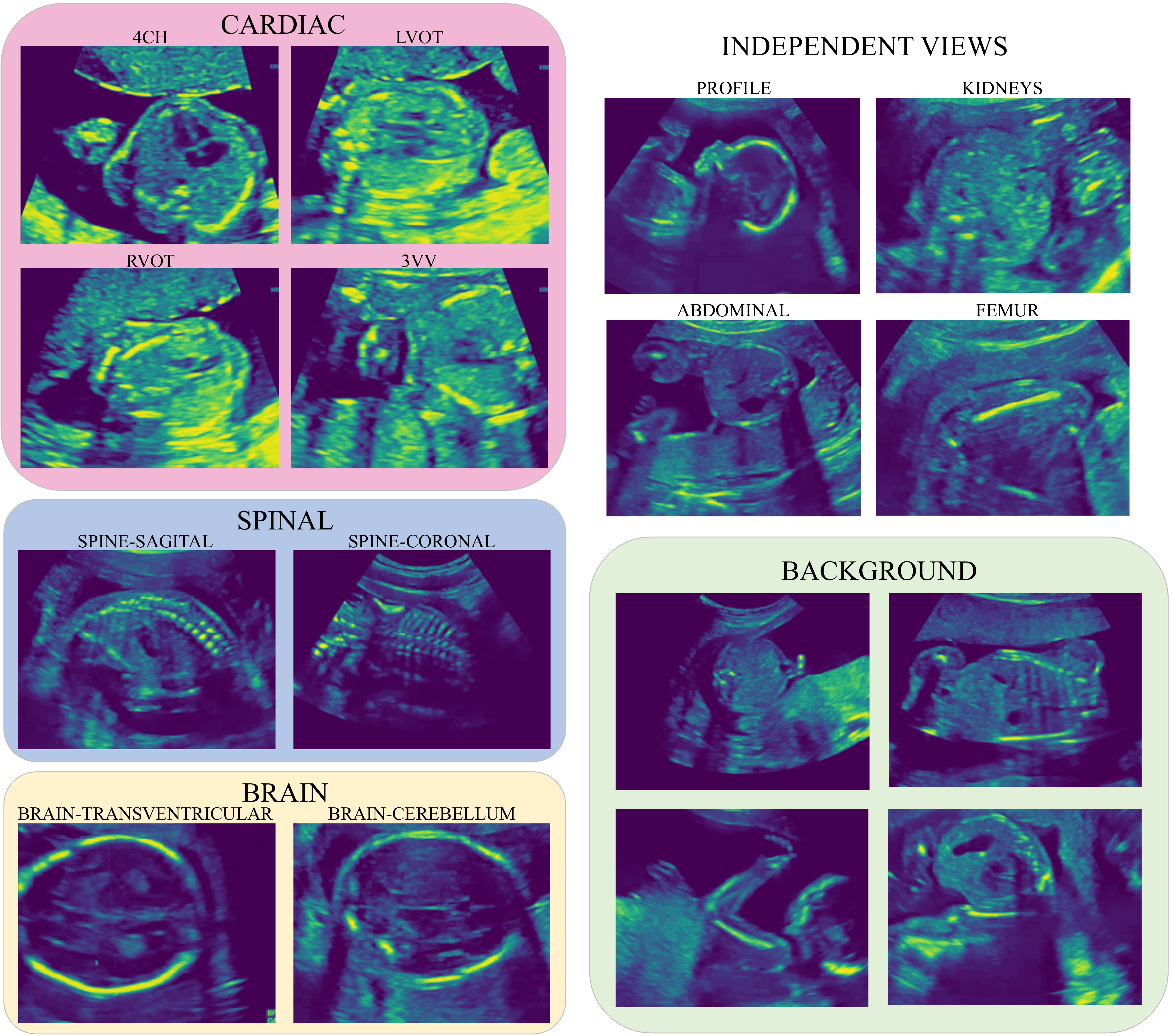}
		\caption{\textit{Standard Plane Views}}
	\end{subfigure}
	\hspace*{\fill}
	
	\caption{\textit{Distribution of the training data (a) and examples of standard views (b).}}
	\label{figures:data_showcase}		
\end{figure*}

\subsection{Dataset}
Our dataset contains 13 anatomical classes plus 1 background class. The entire training dataset consists of 22757 images (class distribution shown in Figure~\ref{figures:data_showcase}.a). For SSL, 100 images are extracted from each class to make up the labelled data $\mathcal{D}_{L}$. The remaining images are treated as unlabelled data $\mathcal{D}_{U}$. Experiments are performed using subsets of $\mathcal{D}_{L}$, specifically using 1, 5, 20, 50, and 100 images per class. The test set follows the same distribution, but totals to 5737 images. Each image is 224x288 pixels. These image frames are derived from a dataset containing 2438 videos from over 2000 volunteers. 

Examples of some of the classes are displayed in Figure~\ref{figures:data_showcase}.b. The cardiac classes are among the most difficult to distinguish. While the spine and brain also span multiple classes, these images are generally more clear and can have significant pose differences (spine).The background class contains a diverse range of images sampled from the videos (excluding frames of standard planes). All extracted samples satisfy a minimum image-space distance between neighbouring frames. This means that most background frames are extracted during rapid probe movement and not when the sonographer slows down to home in on the standard planes. However, some images that resemble standard planes still make it through this na\"ive filtering approach. Also, the background class is larger than any anatomical class (Figure~\ref{figures:data_showcase}.a). In short, the background class introduces class imbalance; is not easily characterized by a single common feature; and contains examples which resemble other classes.   

\subsection{Evaluation}
Fully supervised methods are evaluated with 1, 5, 20, 50, and 100 labelled images per class. Another experiment is performed using the entire labelled training set (total 22757 images). The SSL framework is applied to the cases of 5, 20, and 50 labelled images per class, where there is a large margin for potential improvement. These cases represent a very feasible labelling task compared to labelling all 22757 images. 

Each evaluation is done with and without the background class. To measure the impact of the background class on anatomical classes, accuracy is reported only for the anatomical classes (not background). Accuracy is also reported with a single merged cardiac class. In this case, any cardiac view that is classified as any of the four cardiac views is considered correct. Comparing the overall and grouped cardiac accuracies gives an indication of whether improvements extend to the cardiac classes.

\section{Results}
\begin{figure*}[h]
	\centering
        \includegraphics[trim={0cm 0cm 0cm 0cm},clip, width=0.65\linewidth]{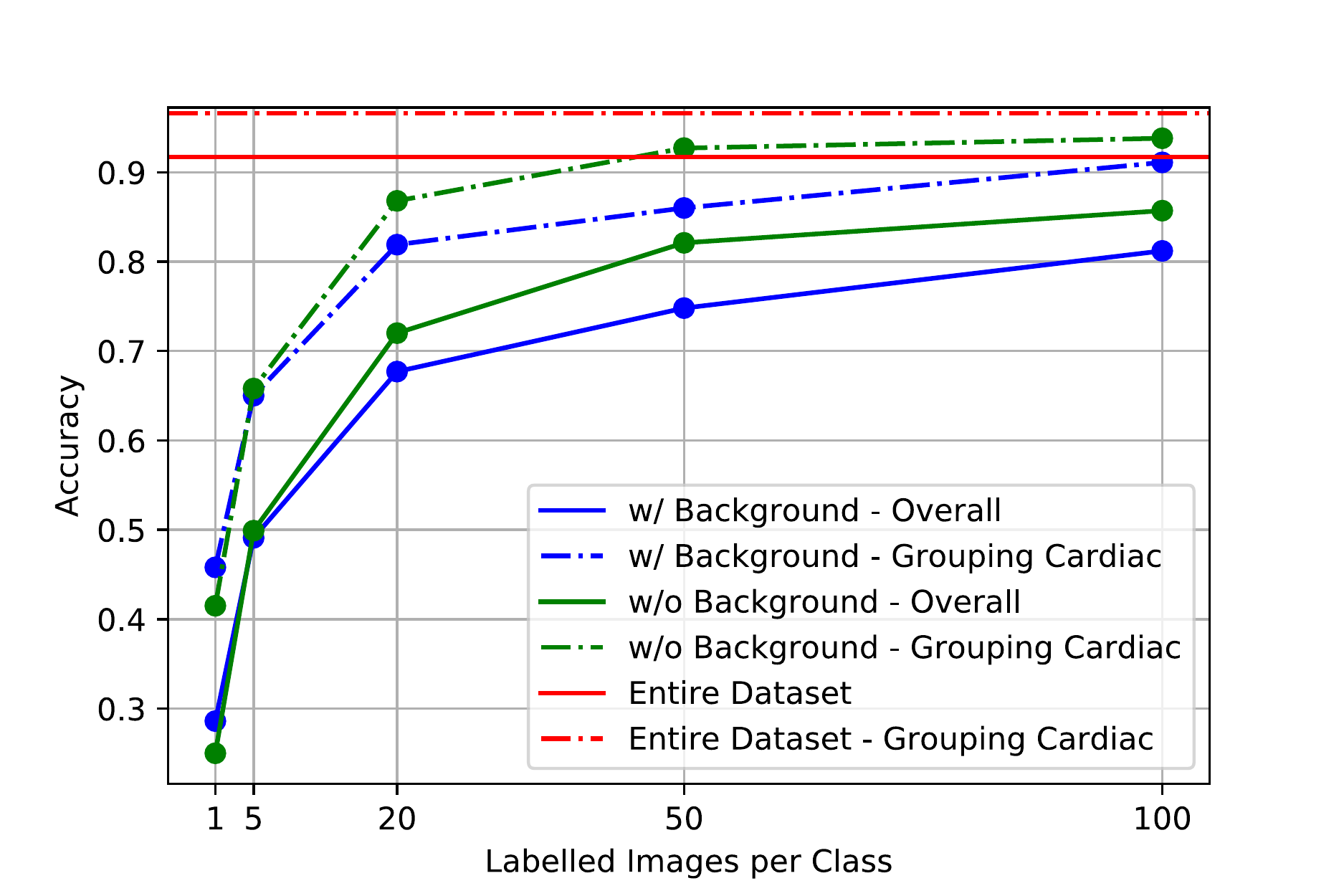}

		\caption{\textit{Supervised baselines with varying number of labelled images per class. Inclusion of the background class increases the difficulty of the classification task. Accuracy is reported as either overall or grouping all cardiac classes into a single class.}}
		\label{figures:sup_acc}
		
\end{figure*}

Accuracy for fully supervised baseline methods is shown in Figure~\ref{figures:sup_acc}. With only 20 labelled examples per class, overall accuracy is near 70\% and grouped cardiac accuracy is over 80\%.All baselines are trained for 50 epochs and did not grossly overfit (except for 1 image per class which was trained for 5 epochs). 

When applying the SSL consistency loss, we find the performance is sensitive to the consistency loss masking threshold, $\eta_{cbm}$. This is particularly true for the cardiac classes. Table~\ref{tab:Thresholding} compares the performance of different thresholds for the cardiac classes. The threshold for all other classes remains constant at 0.75. A cardiac threshold of 0.25 gives best performance and is used for all further experiments.

\begin{table}[ht]
    \centering
    \caption{Confidence mask threshold values for cardiac classes. These experiments use 20 labelled images per class and exclude the background class.}
    \def\arraystretch{1.1}
    \begin{tabular}{c|c|c|c} 
      \textbf{Method} & \def\arraystretch{1} \barr \textbf{Cardiac Confidence}\\ \textbf{Mask Threshold} \earr & \textbf{Grouping Cardiac} & \textbf{Overall}\\
      
      \hline
      \multicolumn{1}{c|}{\multirow{1}{*}{Supervised}} & N/A & \textbf{0.868} & \textbf{0.720}\\
      \hline
      
      \multicolumn{1}{c|}{\multirow{5}{*}{Basic UDA}} & $\eta_{cbm}$ & 0.849 & 0.665\\
      \multicolumn{1}{l|}{} & $\frac{1}{2}\cdot\eta_{cbm}$ & 0.840 & 0.661 \\
      \multicolumn{1}{l|}{} & $\frac{1}{3}\cdot\eta_{cbm}$ & \textbf{0.905} & \textbf{0.728} \\
      \multicolumn{1}{l|}{} & $\frac{1}{4}\cdot\eta_{cbm}$ & 0.831 & 0.664\\
      \multicolumn{1}{l|}{} & $>$ 1 (disabled) & 0.631 & 0.631 \\
      \hline
      
      \multicolumn{1}{c|}{\multirow{2}{*}{\def\arraystretch{1} \barr {UDA Best}\\ {Configuration} \earr }} & $\eta_{cbm}$ & 0.915 & 0.720\\
      \multicolumn{1}{l|}{} & $\frac{1}{3}\cdot\eta_{cbm}$ & \textbf{0.936} & \textbf{0.754} \\
      
    \end{tabular}%
    \label{tab:Thresholding}

\end{table}%

\begin{table}[ht]
    \centering
    \caption{Different configurations for the case of 20 labelled images per class without the background class.}
    \begin{tabular}{c|c|c|c|c} 

      \textbf{Method} & \textbf{Optimizer} & \textbf{TSA Schedule} & \textbf{Grouping Cardiac} & \textbf{Overall}\\
      
      \hline
      \multicolumn{1}{c|}{\multirow{1}{*}{Supervised}} & Adam: 1E-3 & N/A & \textbf{0.868} & \textbf{0.720}\\
      \hline
      
      \multicolumn{1}{c|}{\multirow{5}{*}{UDA}} & Adam: 1E-3 & Linear & 0.905 & 0.728 \\
      
      \multicolumn{1}{l|}{} & Momentum: 1E-3 & Linear & 0.891 & 0.692 \\
      
      \multicolumn{1}{l|}{} & SGD cyclic:[7E-3,5E-2] & Linear &  0.921 & 0.735 \\ 
      
      \multicolumn{1}{l|}{} & Adam: 1E-3 & Log & \textbf{0.936} & \textbf{0.754} \\
      
      \multicolumn{1}{l|}{} & SGD cyclic:[7E-3,5E-2] & Log & 0.935 & 0.744 \\
      
    \end{tabular}
    \label{tab:configurations}
\end{table}

Further experiments are performed with different UDA settings to find the optimal configuration (Table~\ref{tab:configurations}). We find that a log TSA schedule gives the best performance. Log schedules are suggested for cases when the network is less likely to quickly overfit\cite{Xie2019UDA}. 

\begin{figure*}[h]
	\centering
        \includegraphics[width=\linewidth]{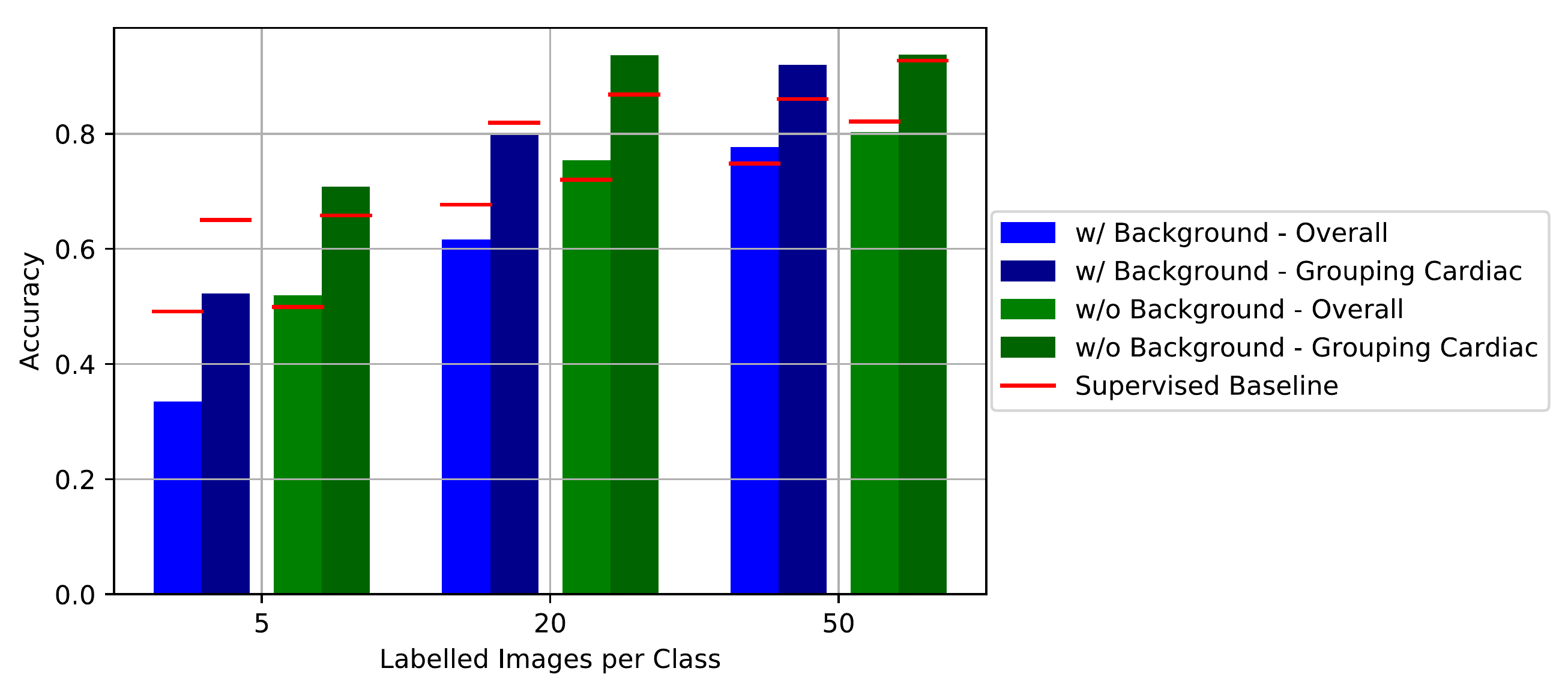}
		\caption{\textit{Overall and grouped cardiac accuracies with and without the background class for 5, 20, and 50 labelled images per class. Red bars indicate supervised baseline performance. Without the background class (green bars), SSL accuracy is almost always above supervised baslines (red). Inclusion of the background class can cause SSL performance to drop below supervised baselines.}}
	\label{figures:bkgd_impact}
\end{figure*}

The best found configuration is then used with 5, 20, and 50 labelled images per class, with and without the background class. Accuracies are reported in Figure~\ref{figures:bkgd_impact}.

\begin{figure*}[h]
	\centering
	\begin{subfigure}[t]{0.32\linewidth}
	
        \includegraphics[width=\linewidth]{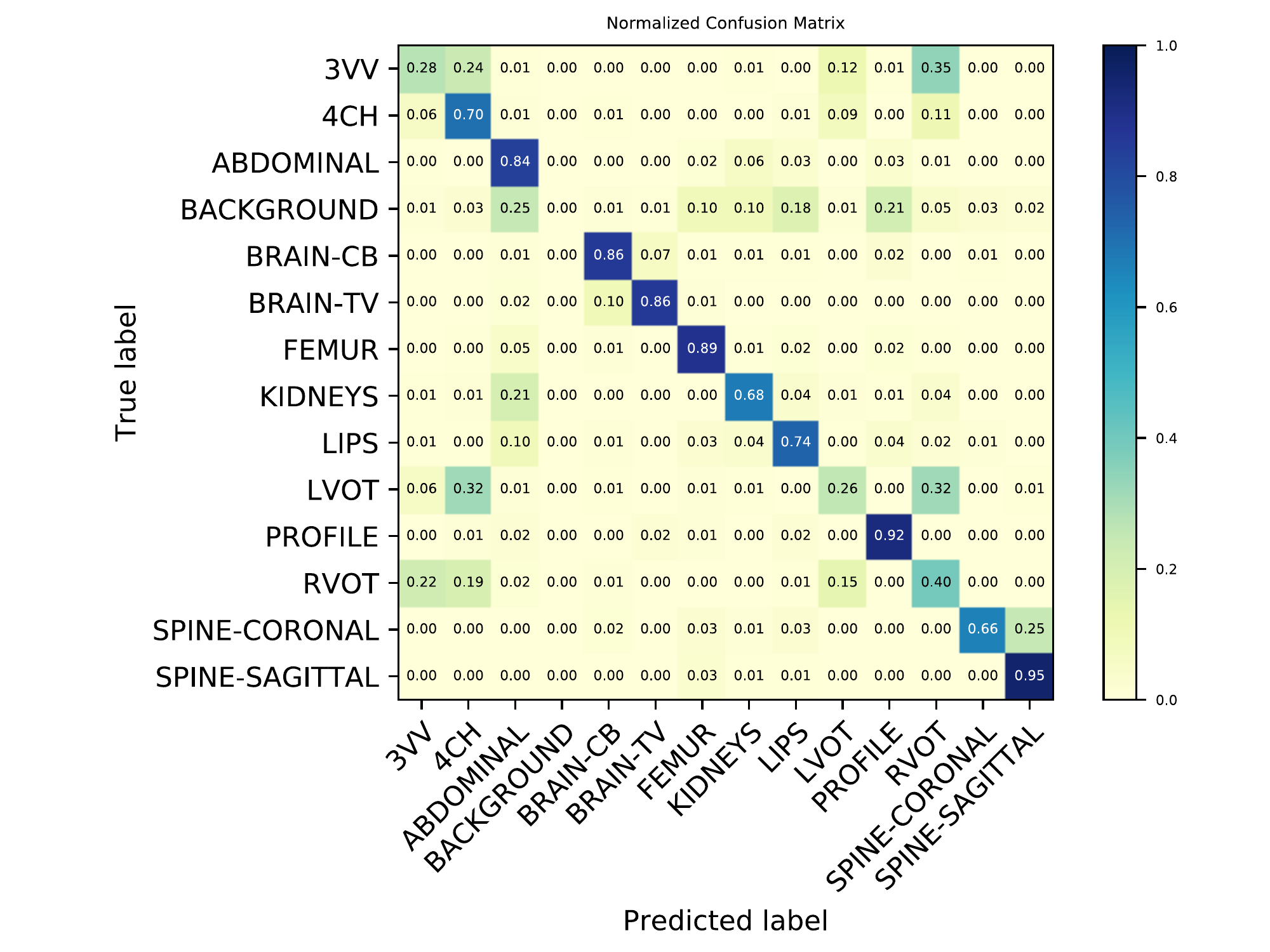}
		\caption{\textit{Supervised baseline}}
	\end{subfigure}
	\begin{subfigure}[t]{0.32\linewidth}
		\includegraphics[trim={0cm 0cm 0cm 0cm},clip,width=\linewidth]{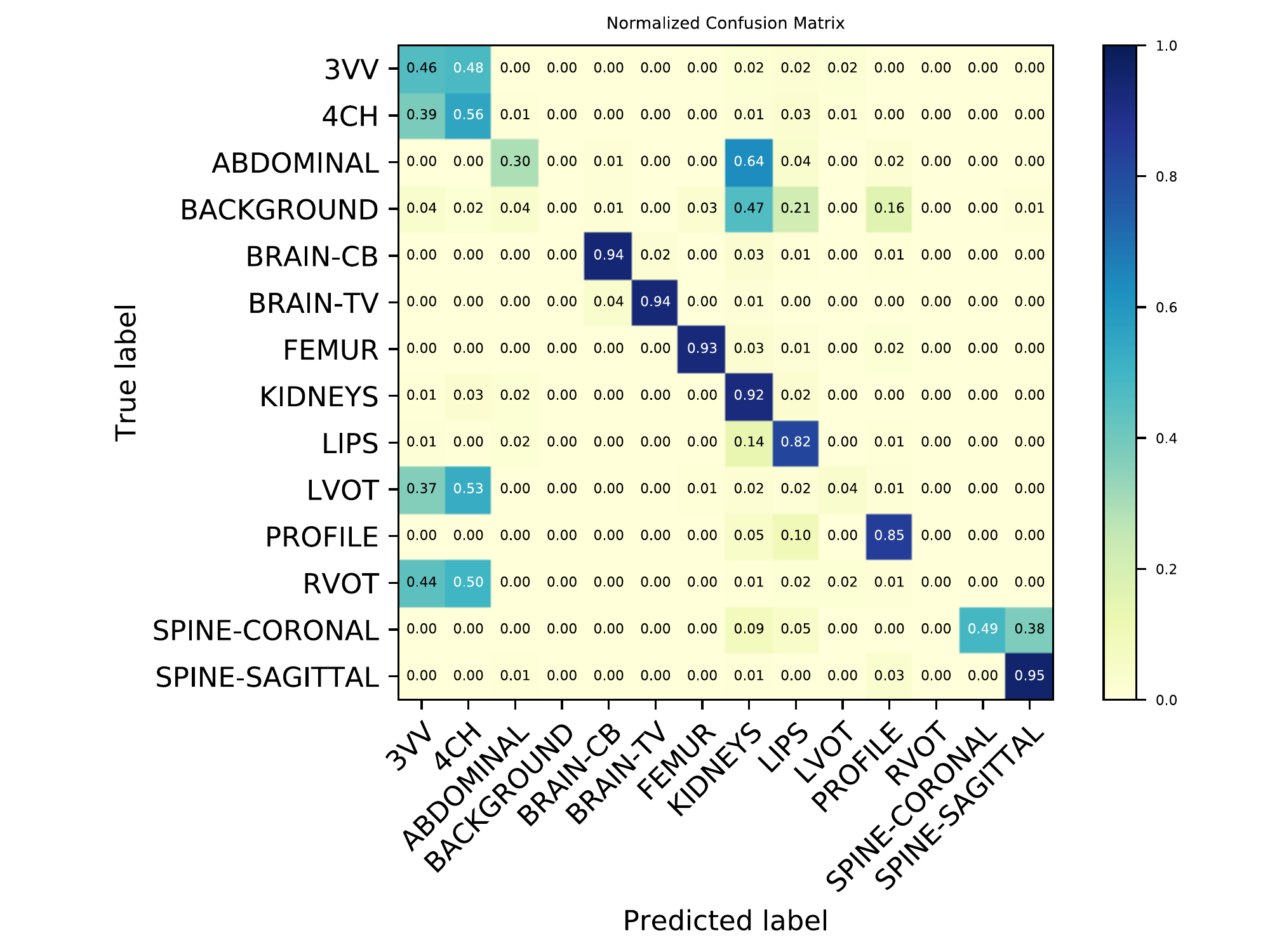}
		\caption{\textit{UDA}}
	\end{subfigure}
	\begin{subfigure}[t]{0.32\linewidth}
		\includegraphics[trim={0cm 0cm 0cm 0cm},clip,width=\linewidth]{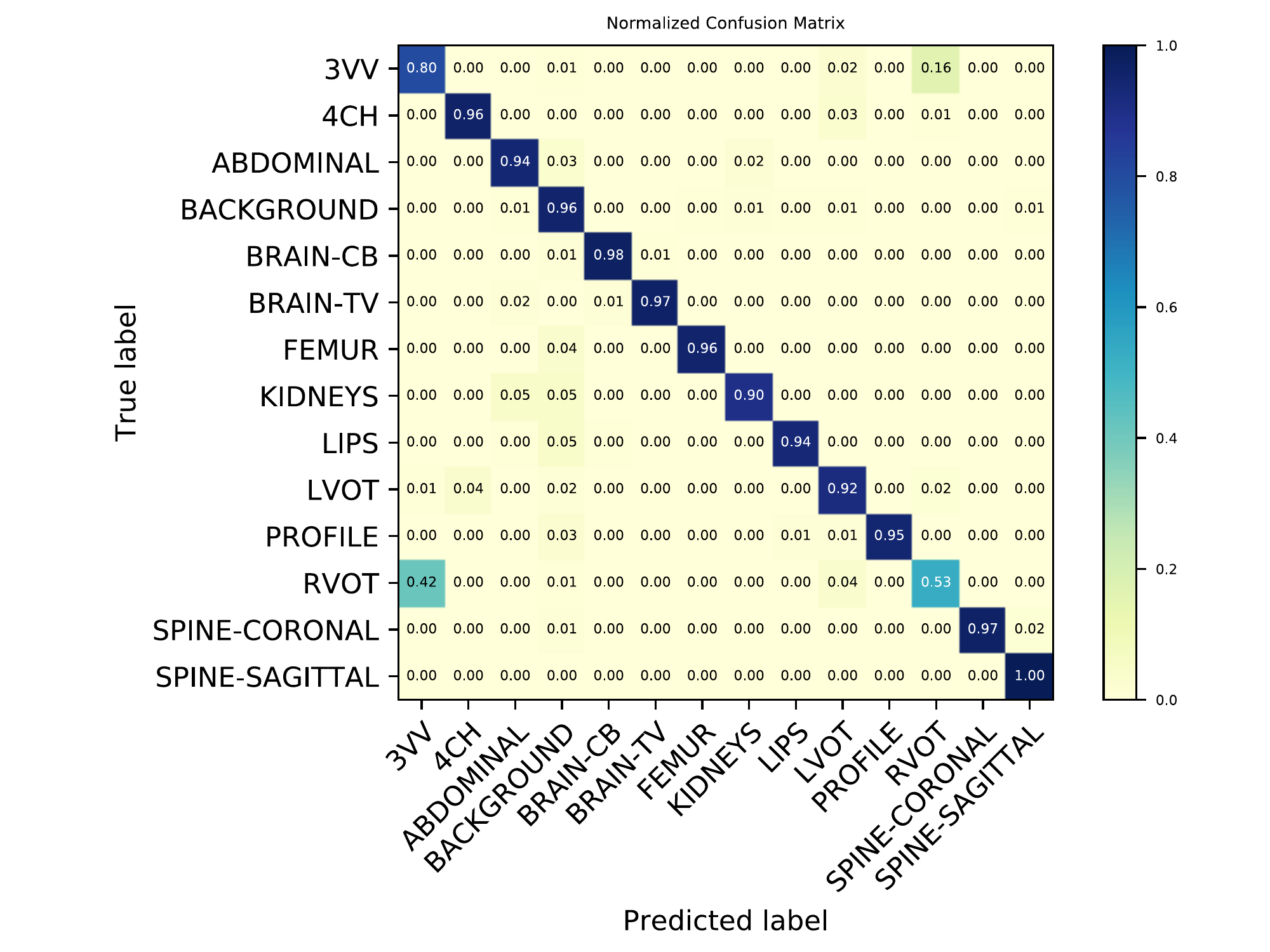}
		\caption{\textit{Supervised on all data}}
	\end{subfigure}
	
	\caption{\textit{Confusion matrices for the case of 20 labelled images per class without background. The supervised baseline (a) performs surprisingly well given the limited data. SSL (b) is able to make considerable improvement for distinct classes, but can increase confusion of cardiac classes. Even when trained on the entire labelled dataset (c), some cardiac classes are prone to confusion.}}
	\label{figures:conf_matrices}		
\end{figure*}

Confusion matrices are displayed in Figure~\ref{figures:conf_matrices}. SSL improves accuracy for distinct classes such as brain, femur, kidney, and lips from mid 0.80 (a - supervised) to mid 0.90 (b - UDA), which approaches the fully supervised performance shown in (c). However, for cardiac classes, confusion is increased when using SSL.

\section{Discussion}
Similarly to Oliver et al.~\cite{Oliver2018RealisticEvaluation}, we show that supervised baselines are surprisingly accurate (Figure~\ref{figures:sup_acc}). There is also a clear diminishing return of increasing the number of labelled images, which starts as early as 20 examples per class. Inclusion of the background class decreases accuracy in almost all cases. This indicates that the background class adds difficulty even in the fully supervised case.

Investigating sensitivity to confidence thresholds (Table~\ref{tab:Thresholding}), we see that $\frac{1}{3}\cdot\eta_{cbm}$ (0.25) is the only setting that improves both grouped cardiac and overall accuracy for the basic UDA implementation. A value near 0.25 is reasonable given that the network must divide its confidence over 4 very similar cardiac classes. Even for the best UDA configuration, a threshold of 0.25 for the cardiac classes makes a considerable difference; without it, overall accuracy does not improve from the supervised baseline. 

Figure~\ref{figures:bkgd_impact} displays that without the background class (green bars), the SSL regime can almost always improve upon the fully supervised baseline. The only exception being the overall accuracy for 50 labelled examples. In this case the supervised baseline is already quite high and any further improvement would likely depend on the cardiac classes which the SSL method struggles with. In contrast, the inclusion of the background class (blue bars), not only reduces the accuracy of the fully supervised baselines, but tends to be harmful to the SSL method. For most of the blue bars, the SSL method fails to match, let alone surpass, the baseline accuracy. This illustrates the negative impact of including confounding images in the unlabelled data. Again, the case with 50 labelled examples is the exception. Perhaps 50 labelled examples is sufficient to capture the majority of the variation in the background class, preventing it from having a negative impact on the SSL loss.

While the SSL has been shown to increase both grouped cardiac and overall accuracies, the confusion matrices in Figure~\ref{figures:conf_matrices} clearly show that these improvements are often at the expense of the cardiac classes. It seems unlabelled data can help the network to learn when the classes are inherently more distinct, but can cause harm when classes are inherently similar.

\section{Conclusion}
Supervised baselines provide surprisingly reliable performance even in extremely low data regimes (e.g. accuracy of 50\% from only 5 labelled images per class). Recent developments in SSL can further improve this performance. However, irregular data can cause SSL to be detrimental rather than beneficial. Also, classes with high similarity, such as cardiac views, can see an increase in confusion. For such classes, injecting domain knowledge (e.g. lowering cardiac confidence thresholds) may be necessary to supplement a lack of labelled examples.


Acknowledgements: Support from Wellcome Trust IEH Award iFind project [102431]. JT is supported by the ICL President's Scholarship. 
%
%
%
%

\begin{thebibliography}{10}
\providecommand{\url}[1]{\texttt{#1}}
\providecommand{\urlprefix}{URL }
\providecommand{\doi}[1]{https://doi.org/#1}

\bibitem{Athiwaratkun2019FastSWA}
Athiwaratkun, B., Finzi, M., Izmailov, P., Wilson, A.G.: There are many
  consistent explanations of unlabeled data: Why you should average. ICLR
  (2019)

\bibitem{Baumgartner2017SonoNet}
Baumgartner, C.F., Kamnitsas, K., Matthew, J., Fletcher, T.P., Smith, S., Koch,
  L.M., Kainz, B., Rueckert, D.: {SonoNet: Real-Time Detection and Localisation
  of Fetal Standard Scan Planes in Freehand Ultrasound}. IEEE Transactions on
  Medical Imaging  \textbf{36}(11),  2204--2215 (2017)

\bibitem{cai2018multi}
Cai, Y., Sharma, H., Chatelain, P., Noble, J.A.: Multi-task sonoeyenet:
  Detection of fetal standardized planes assisted by generated sonographer
  attention maps. In: International Conference on Medical Image Computing and
  Computer-Assisted Intervention. pp. 871--879 (2018)

\bibitem{Chen2015PlaneDetectionDomainTransfer}
Chen, H., Ni, D., Qin, J., Li, S., Yang, X., Wang, T., Heng, P.A.: {Standard
  Plane Localization in Fetal Ultrasound via Domain Transferred Deep Neural
  Networks}. IEEE Journal of Biomedical and Health Informatics  \textbf{19}
  (2015)

\bibitem{Cheplygina2019NotSoSupervised}
Cheplygina, V., de~Bruijne, M., Pluim, J.P.: Not-so-supervised: A survey of
  semi-supervised, multi-instance, and transfer learning in medical image
  analysis. Medical Image Analysis  \textbf{54},  280 -- 296 (2019)

\bibitem{Cubuk2019AutoAugment}
Cubuk, E.D., Zoph, B., Mane, D., Vasudevan, V., Le, Q.V.: {AutoAugment}:
  Learning augmentation policies from data. CVPR  (2019)

\bibitem{Grandvalet2005EntMin}
Grandvalet, Y., Bengio, Y.: Semi-supervised learning by entropy minimization.
  NeurIPS  (2005)

\bibitem{Kong2018PlaneDetectionMultiScale}
Kong, P., Ni, D., Chen, S., Wang, T., Lei, B.: {Automatic and Efficient
  Standard Plane Recognition in Fetal Ultrasound Images via Multi-scale Dense
  Networks}. Data Driven Treatment Response Assessment and Preterm, Perinatal,
  and Paediatric Image Analysis, LNCS  \textbf{11076} (2018)

\bibitem{Laine2017Pi}
Laine, S., Aila, T.: Temporal ensembling for semi-supervised learning. ICLR
  (2017)

\bibitem{Oliver2018RealisticEvaluation}
Oliver, A., Odena, A., Raffel, C., Cubuk, E.D., Goodfellow, I.J.: Realistic
  evaluation of deep semi-supervised learning algorithms. NeurIPS  (2018)

\bibitem{Singh2008HelpDoesnt}
Singh, A., Nowak, R., Zhu, J.: Unlabeled data: Now it helps, now it doesn't.
  NeurIPS  (2008)

\bibitem{Verma2019ICT}
Verma, V., Lamb, A., Kannala, J., Bengio, Y.: Interpolation consistency
  training for semi-supervised learning. IJCAI  (2019)

\bibitem{WHO2016recommendations}
{World Health Organization}, et~al.: WHO recommendations on antenatal care for
  a positive pregnancy experience. World Health Organization (2016)

\bibitem{Xie2019UDA}
Xie, Q., Dai, Z., Hovy, E., Luong, M.T., Le, Q.V.: Unsupervised data
  augmentation for consistency training. arXiv preprint arXiv:1904.12848
  (2019)

\end{thebibliography}

\end{document}